\documentclass[journal]{IEEEtran}
\usepackage{graphicx}
\usepackage{amstext}
\usepackage{mathrsfs}
\usepackage{amssymb}
\usepackage{amsmath}
\usepackage{bm}
\allowdisplaybreaks[4]
\usepackage{epstopdf}
\usepackage{multicol}
\usepackage{stfloats}
\usepackage{url}
\usepackage{color}
\usepackage{enumerate}
\usepackage{breqn}
\usepackage{bbm}
\usepackage{cite}
\usepackage{caption}
\usepackage{enumitem}
\usepackage{array}
\usepackage{subcaption}
\usepackage{multirow}
\usepackage{booktabs}
\usepackage{bookmark}
\usepackage[ruled,vlined]{algorithm2e}
\usepackage{booktabs}
\newcommand{\ie}{\textit{i.e., }}
\newcommand{\eg}{\textit{e.g., }}
\usepackage[normalem]{ulem}
\useunder{\uline}{\ul}{}
\newcommand{\etal}{\textit{et al.}}
\newcommand{\re}[1]{{\color{black}#1}}

\ifCLASSINFOpdf
\else
\fi
\begin{document}

\title{Towards Efficiently Evaluating the Robustness of Deep Neural Networks in IoT Systems: A GAN-based Method}
\author{Tao Bai,
        Jun Zhao,~\IEEEmembership{Member,~IEEE,}
        Jinlin Zhu, 
        Shoudong Han,
        Jiefeng Chen,
        Bo Li,
        Alex Kot,~\IEEEmembership{Fellow,~IEEE}

\thanks{T. Bai and J. Zhao are with the School of Computer Science and Engineering, Nanyang Technological University, Singapore 639798 (Email: bait0002@ntu.edu.sg; junzhao@ntu.edu.sg).}
\thanks{J. Zhu and A. Kot are with the School of Electrical and Electronic Engineering, Nanyang Technological University, Singapore 639798 (Email: jerry.zhu@ntu.edu.sg; eackot@ntu.edu.sg). }
\thanks{S. Han is with the National Key Laboratory of Science and Technology on Multispectral Information Processing, and School of Artificial Intelligence and Automation at Huazhong University of Science and Technology (HUST), Wuhan, China. (Email: shoudonghan@hust.edu.cn).}
\thanks{J. Chen is with the Department of Computer Sciences, University of Wisconsin-Madison, 1210 W Dayton St, Madison WI 53706-1613, USA (Email: jchen662@wisc.edu).}
\thanks{B. Li is with Computer Science Department, University of Illinois at Urbana-Champaign, 4310 Siebel Center 201 N. Goodwin Ave. Urbana, IL 61801, USA (Email: lbo@illinois.edu).}
}

\maketitle

\begin{abstract}
Intelligent Internet of Things (IoT) systems based on deep neural networks (DNNs) have been widely deployed in the real world.
However, DNNs are found to be vulnerable to adversarial examples, which raises people's concerns about intelligent IoT systems' reliability and security.
Testing and evaluating the robustness of IoT systems becomes necessary and essential.
Recently various attacks and strategies have been proposed, but the efficiency problem remains unsolved properly.
Existing methods are either computationally extensive or time-consuming, which is not applicable in practice.
In this paper, we propose a novel framework called Attack-Inspired GAN (AI-GAN) to generate adversarial examples conditionally.
Once trained, it can generate adversarial perturbations efficiently given input images and target classes.
We apply AI-GAN on different datasets in white-box settings, black-box settings and targeted models protected by state-of-the-art defenses. 
Through extensive experiments, AI-GAN achieves high attack success rates, outperforming existing methods, and reduces generation time significantly.
Moreover, for the first time, AI-GAN successfully scales to complex datasets \eg CIFAR-100 and ImageNet, with about $90\%$ success rates among all classes.
\end{abstract}


\IEEEpeerreviewmaketitle

\section{Introduction}
\IEEEPARstart{D}{eep} neural networks have achieved great success in the last few years and drawn tremendous attention from both academia and industry.
Nowadays, it has been intensively applied in smart industrial or daily-use IoT systems and devices, including those safety and security-critical ones, such as autopilots~\cite{xu2018analyzing,yu2018security}, face recognition on mobile devices~\cite{ijiri2006security,pan2017future}, traffic transportation systems~\cite{zanella2014internet,li2018intelligent} and intelligent manufacturing~\cite{tang2019reconfigurable,li2018service}. 
With the rapid development and deployment of these intelligent IoT systems, safety concerns rise from society.

Recent studies have found that deep neural networks used by IoT systems are vulnerable to adversarial examples~\cite{DBLP:journals/corr/SzegedyZSBEGF13,DBLP:journals/corr/GoodfellowSS14,9174940,ding2020towards}. 
Adversarial examples are usually crafted by adding carefully-designed imperceptible perturbations on legitimate samples.
In human's eyes, the appearances of adversarial examples are the same as their legitimate copies, while the predictions from deep learning models are different.
The existence of adversarial examples has dramatically challenged smart IoT systems' safety and reliability.
Thus the importance and necessity of efficiently evaluating the reliability of safety-critical systems are getting increasingly high~\cite{8454402}.

Many researchers have managed to evaluate the robustness of deep neural networks in different ways, such as box-constrained \mbox{L-BFGS}~\cite{DBLP:journals/corr/SzegedyZSBEGF13}, Fast Gradient Sign Method (FGSM)~\cite{DBLP:journals/corr/GoodfellowSS14}, Jacobian-based Saliency Map Attack (JSMA)~\cite{papernot2016limitations}, C\&W attack~\cite{carlini2017towards} and Projected Gradient Descent~(PGD) attack~\cite{DBLP:conf/iclr/MadryMSTV18}.
These attack methods are optimization-based with proper distance metrics $L_{0}$, $L_{2}$ and $L_{\infty}$ to restrict the magnitudes of  perturbations and make the presented adversarial examples visually natural.
These methods are based on optimization, which is usually time-consuming, computation-intensive, and need to access the target models at the inference period for strong attacks. 
These properties make optimization-based methods not applicable to test IoT systems in practice.

Some researchers employ generative models \eg GAN~\cite{goodfellow2014generative} to produce adversarial perturbations~\cite{XiaoLZHLS18,DBLP:conf/cvpr/PoursaeedKGB18}, or generate adversarial examples directly~\cite{NIPS2018_8052}. 
Compared to optimization-based methods, generative models significantly  reduce the time of adversarial examples generation.
Yet, existing methods have two apparent limitations: 
1) The generation ability is limited \ie they can only perform one specific targeted attack at a time. 
Re-training is needed for different targets.
2) they can hardly scale to real world images. 
Most prior works evaluated their methods only on MNIST and CIFAR-10, which is not feasible for complicated reality tasks.

To solve the above problems, we propose a new variant of GAN to generate adversarial perturbations conditionally and efficiently, which is named \mbox{Attack-Inspired} GAN (\mbox{AI-GAN}) and shown in Fig.~\ref{fig:overall_arc}:
a generator is trained to perform targeted attacks with clean images and targeted classes as inputs;
a discriminator with an auxiliary classifier for classification generated samples in addition to discrimination.
Unlike existing works, we propose to add an attacker and train the discriminator adversarially.
On the one hand, the discriminator after adversarial training gets adversarially robust.
This robust discriminator enhances the generator's attack abilities.
On the other hand, a robust discriminator can also stabilize the GAN training process~\cite{zhou2018dont,Liu_2019_CVPR}.
Inspired by other complicated tasks in deep learning \eg object detection, which usually adopts pre-trained models as backbones, we use a pre-trained model in our generator to compress data for better scalability.
On evaluation, we mainly select four datasets with different classes and image sizes and compare our approach with four representative methods in white-box settings, black-box settings and under defences.
From the experiments, we conclude that 1) our model and loss function are useful, with much-improved efficiency and scalability;
2) our approach generates comparable or even stronger attacks (for the most time) than existing methods under the same $L_{\infty}$ bound of perturbations in various experimental settings. 

We summarize our contributions as follows: 
\begin{enumerate}
\item Unlike existing methods, we propose a novel framework called \mbox{AI-GAN} where a generator, a discriminator, and an attacker are trained jointly. 
\item To our best knowledge, AI-GAN is the first GAN-based approach to generate perceptually realistic adversarial examples given inputs and targets, and scales to complicated datasets \eg CIFAR-100 and ImageNet, achieving high attack success rates~($\approx 90\%$).
\item Through extensive experiments, \mbox{AI-GAN} shows strong attack abilities, outperforming existing methods in both white-box and black-box settings, and saves time significantly.
\item We show AI-GAN achieves comparable or higher success rates on target models protected by the state-of-the-art defense methods.
\end{enumerate}

The remainder of this paper is organized as follows:
In Section~\ref{sec:Related Work}, we briefly review the literature related to adversarial examples and generative models.
In Section~\ref{sec:preli}, we introduce some representative attacks.
Then Section~\ref{sec:approach} elaborates our proposed method for generating adversarial examples efficiently.
In Section~\ref{sec:experiment}, we show the experimental results on various datasets with different settings.
In Section~\ref{sec:disc}, we discuss the efficiency of attack generation.
Finally, Section~\ref{sec:con} concludes the paper.

\section{Related Work}\label{sec:Related Work}

\subsection{Adversarial Examples} 
Adversarial examples, which are able to mislead deep neural networks, are first discovered by~\cite{DBLP:journals/corr/SzegedyZSBEGF13}.
They manage to maximize the network's prediction error by adding hardly perceptible perturbations to benign images.
Since then, various attack methods have been proposed.  
\cite{DBLP:journals/corr/GoodfellowSS14} developed Fast Gradient Sign Method (FGSM) to compute the perturbations efficiently using back-propagation.
The perturbations could be expressed as 
$\eta=\epsilon \operatorname{sign}\left(\nabla_{x} J(\boldsymbol{\theta}, x, y)\right)$,
where $J(\theta, x, y)$ 
represents the cross entropy loss function, and $\epsilon$ is a constant. 
Thus adversarial examples are expressed as $x_{A}=x+\eta$.
One intuitive extension of FGSM is Basic Iterative Method~\cite{DBLP:journals/corr/KurakinGB16}
which executes FGSM many times with smaller $\epsilon$.
\cite{papernot2016limitations} proposed Jacobian-based Saliency Map Attack (JSMA)
with $L_{0}$ distance.
The saliency map discloses the likelihood of fooling the target network when modifying 
pixels in original images.
Optimization-based methods have been proposed to generate quasi-imperceptible 
perturbations with constraints in different distance metrics.
\cite{carlini2017towards} designed a set of attack methods.
The objective function to be minimized is $\|\delta\|_{p}+c \cdot f(x+\delta)$,
where $c$ is an constant and $p$ could be 0, 2 or $\infty$.
\cite{Chen2017} also proposed an algorithm approximating the gradients
of targeted models based on Zeroth Order Optimization (ZOO).
\cite{DBLP:conf/iclr/MadryMSTV18} introduced a convex optimization
method called Projected Gradient Descent (PGD) to generate adversarial examples, which is 
proved to be the strongest first-order attack.
However, such methods are usually time-consuming because the optimization process is slow,
and they are only able to generate perturbations once a time.

\subsection{Generative Adversarial Networks (GANs)}
GAN is firstly proposed by~\cite{goodfellow2014generative} 
and has achieved great success in various tasks \cite{isola2017image,zhu2017unpaired,reed2016generative}. 
GAN consists of two competing neural networks with different objectives, called
the discriminator $D(x)$ and the generator $G(z)$ respectively.
The training phase could be seen as a \mbox{min-max}  game of the discriminator and the generator.
The generator is trained to generate fake images to fool the discriminator 
while the discriminator tries to classify the real images and the fake images.
Usually the generator and the discriminator are trained alternatively while the other one
is fixed. 
The original objective function is denoted as 
\begin{equation}
\begin{aligned}
\min_{G} \max_{D} V(D, G)=& \mathbb{E}_{x \sim p_{\text {data }}(x)}[\log D(x)] \\ &  +\mathbb{E}_{z \sim {p_{z}}(z)}[\log (1-D(G(z)))]
\end{aligned}
\end{equation} 
However, the training of vanilla GAN is very unstable. 
To stabilize the GAN training, a variety of improvements of GAN are developed such as DCGAN~\cite{radford2015unsupervised}, 
WGAN~\cite{arjovsky2017wasserstein} and WGAN with gradient penalty~\cite{gulrajani2017improved}.
In this paper, we show that a robust discriminator is good for stable training.

\subsection{Generating Adversarial Examples via Generative Models} 
Generative models are usually used to create new data  because of their powerful representation ability.
\cite{DBLP:conf/cvpr/PoursaeedKGB18} firstly applied generative models to generate four types of adversarial perturbations (universal or image dependent, targeted or non-targeted) with U-Net~\cite{ronneberger2015u} and ResNet~\cite{he2016deep} architectures. 
\cite{mao2020gap++} extended the idea of~\cite{DBLP:conf/cvpr/PoursaeedKGB18} with conditional generation.
\cite{XiaoLZHLS18} used the idea of GAN to make adversarial examples more realistic.
Different from the above methods generating adversarial perturbations, some other methods generate adversarial examples directly, which are called unrestricted adversarial examples~\cite{NIPS2018_8052}.
Song\textit{~et~al.~}\cite{NIPS2018_8052} proposed to search the latent space of pretrained ACGAN~\cite{odena2017conditional} to find adversarial examples.
Note that all these methods are only evaluated on simple datasets \eg MNIST and CIFAR-10.

\section{Preliminaries}\label{sec:preli}
In this section, we introduce some common adversarial attacks. 
\re{Given an input image $x \in \mathbb{R}^{h\times w \times c}$, the adversary aims to find a perturbation $\delta$ so that $f(x+\delta) \neq f(x)$.
Due to the intractability of $f(x+\delta) \neq f(x)$ in optimization, the objective of the adversary is to maximize the loss function $J$ of the target classifier $f$.
The norm of $\delta$ is usually restricted by a small scalar value $\epsilon$, and $L_{\infty}$ is a popular norm used in the literature~\cite{DBLP:conf/iclr/MadryMSTV18}.}

\subsection{Fast Gradient Sign Method (FGSM)}
\re{
Goodfellow~\etal~\cite{DBLP:journals/corr/GoodfellowSS14} hypothesized the ``linearity" of deep network models in high dimensional spaces that small perturbations on inputs would be exaggerated by the deep models.
FGSM exploits such linearity to generate adversarial examples.
Specifically, FGSM computes the sign of gradients with respect to the input $x$, and adversarial perturbation $\delta$ is expressed as:
\begin{equation}
\centering
\begin{aligned}
\delta = \varepsilon \operatorname{sign}(\nabla J_{x}(\theta, x, y)), 
\end{aligned}
\end{equation}
where $\theta$ represents the parameters of the targeted classifier, and $J(\theta, x, y)$ denotes the loss function, and $\epsilon$ is the magnitude of perturbations.
}
\subsection{C\&W Attack}
\re{C\&W Attack proposed by~\cite{carlini2017towards} is designed to generate quasi-imperceptible adversarial examples with a high attack ability.
The objective of C\&W Attack is formulated as follows:
\begin{equation}
\operatorname{minimize} \quad c \cdot g(x+\delta)+\|\delta\|_{\infty}
\label{eqn:cw}
\end{equation}
with $g(x)$ defined as
\begin{equation}
g\left(x\right)=\max \left(\max_{i \neq y} \left\{f_{i}\left(x\right) -f_{y}\left(x\right)\right\},-\kappa\right)
\end{equation}
where $c$ and $\kappa (\geq 0)$ are hyper-parameters, $y$ is correct class $x$ belongs to, and $f_{i}\left(x\right)$ is the probability of class $i$.
Thus, the first term in Equation~\ref{eqn:cw} is to find the worst samples, while the second term is used to constrain the magnitude of the perturbations.
}
\subsection{Projected Gradient Descent (PGD) Attack}
\re{PGD attack is a multi-step variant of FGSM with a smaller step size.
To make sure that adversarial examples are in a valid range, PGD attack projects the adversarial examples into the neighbor $\mathcal{S}$ of the original samples after every update, which is defined by $\epsilon$ and $l_{\infty}$.
Formally, the update scheme is expressed as
\begin{equation}
x^{t+1}=\Pi_{x+\mathcal{S}(x,\varepsilon)}\left(x^{t}+\alpha \operatorname{sgn}\left(\nabla_{x} J(\theta, x, y)\right)\right),
\end{equation}
where $\mathcal{S}(x,\varepsilon):=\left\{x+\delta \in \mathbb{R}^{h\times w \times c}  \mid  \|\delta\| _{p} \leq \varepsilon\right\}$ is the set of allowed perturbations to ensure the validity of $x^{t+1}$ and $\alpha$ is the step size.}

\subsection{AdvGAN}
\begin{figure}[htb!]
\centering
  \includegraphics[width=\columnwidth]{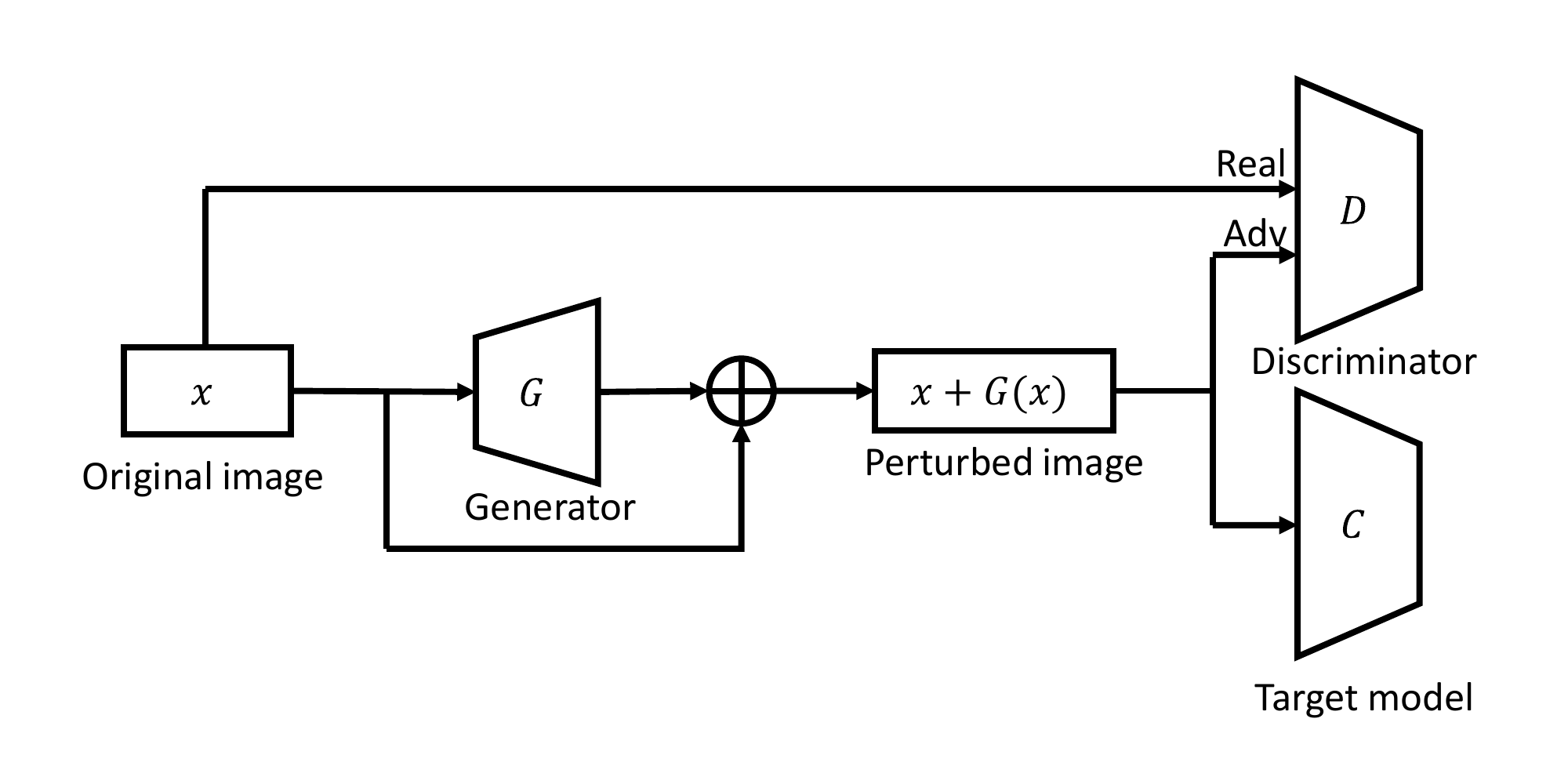}
\caption{The architecture of AdvGAN.}
\label{fig:advgan}
\end{figure}

AdvGAN is the first work which employs GAN to generate targeted adversarial examples.
\re{The architecture of AdvGAN is illustrated in Fig.~\ref{fig:advgan}.
As we can see, AdvGAN consists of a generator, a discriminator, and a pretrained target model $C$.
Given an input $x$, the goal of AdvGAN is to generate adversarial perturbations $\delta$ so that $C(x+\delta) = t$, where $t$ is the target class pre-defined by the adversary and $x$ doesn't belong to class $t$.
}

\section{Our Approach}\label{sec:approach}
\subsection{Problem Definition}
Consider a classification network $f$ trained on dataset $\mathcal{X} \subseteq \mathcal{R}^n$, with $n$ being the dimension of inputs.
And suppose $(x_i, y_i)$ is the $i^{th}$ instance in the training data, where $x_i \in \mathcal{X}$ is generated from some unknown distribution $\mathcal{P}_{\mathrm{data}}$, and $y_i \in \mathcal{Y}$ is the ground truth label.
The classifier $f$ is trained on natural images and achieves high accuracy.
The goal of an adversary is to generate an adversarial example $x_{adv}$, which can fool $f$ to output a wrong prediction and looks similar to $x$ in terms of some distance metrics.
We use $L_{\infty}$ to bound the magnitude of perturbations.
\re{There are two types of such attacks: untargeted attacks and targeted attacks. 
Specifically, untargeted attacks are the attacks that the adversary uses to mislead the target model to predict any of the incorrect classes, while targeted attacks are the attacks that the adversary uses to mislead the target model to a pre-defined class except the true class. 
Given an instance $(x, y)$, for example, the adversarial example generated by the adversary is $x_{adv}$.
If the goal of the adversary is $f(x_{adv}) \neq y$, this is untargeted attack; or the goal of the adversary is $f(x_{adv}) = t$, where $t$ is the target class defined by the adversary and $t \neq y$, this is targeted attack.
As claimed in~\cite{pmlr-v80-athalye18a}, misclassification caused by untargeted attacks sometimes may not be meaningful for closely related classes; \eg a German shepherd classified as a Doberman.
They suggested that targeted attacks are more recommended for evaluation. 
In addition, generating targeted attacks is strictly harder than developing untargeted attacks~\cite{pmlr-v80-athalye18a}.
Thus, we mainly focus on targeted attacks in this paper.
}

\subsection{Proposed Framework}
\begin{figure*}[htb!]
\centering
  \includegraphics[width=\textwidth]{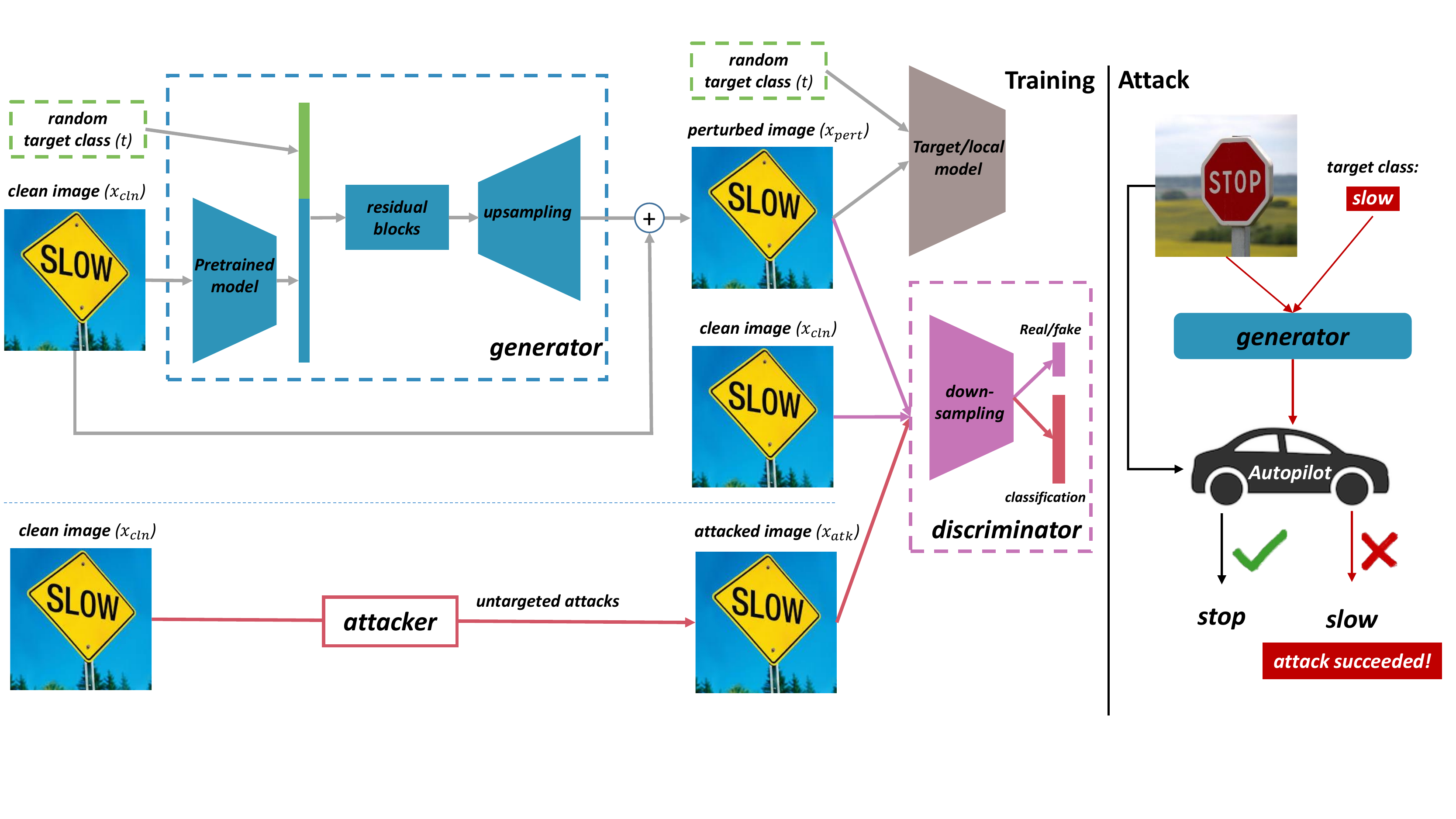}
\caption{\re{Overview of our approach. 
\textit{Left}: the Training Scheme of AI-GAN. See Section IV-B for the detailed explanation.
\textit{Right}: an Illustration of Targeted Attacks by AI-GAN. Given the ``stop sign'' image and the target class ``slow'', the perturbed image is misclassified in the ``slow'' class (red line). Without the modification by AI-GAN, this ``stop sign'' image is classified correctly (black line).
}
}
\label{fig:overall_arc}
\end{figure*}

We propose a new variant of conditional GAN called Attack-Inspired GAN (\mbox{AI-GAN}) to generate adversarial examples conditionally and efficiently.
As shown in Fig.~\ref{fig:overall_arc}, the overall architecture of \mbox{AI-GAN} consists of a generator $G$, a two-head discriminator $D$, an attacker $A$, and a target classifier $C$.
Within our approach, both generator and discriminator are trained in an end-to-end way:
the generator generates and feeds perturbed images to the discriminator.
Meanwhile, clean images sampled from training data and their attacked copies are fed into the discriminator as well.
Specifically, the generator $G$ takes a clean image \re{$x_{cln}$} and the target class label $t$ as inputs to generate adversarial perturbations \re{$G(x_{cln}, t)$}.
$t$ is sampled randomly from the dataset classes except the correct class.
An adversarial example \re{$x_{pert}:=x_{cln} + G(x_{cln}, t)$} can be obtained and sent to the discriminator $D$.
Other than the adversarial examples generated by $G$, the discriminator $D$ also takes the clean images \re{$x_{cln}$} and \re{untargeted adversarial examples $x_{atk}$} generated by the attacker $A$.
So $D$ not only discriminates clean/perturbed images, but also classifies adversarial examples correctly.

\textbf{Discriminator.}
The discriminator is originally designed for classifying real or fake images, forcing the generator to generate convincing images~\cite{goodfellow2014generative,XiaoLZHLS18}.
However, it is known that forcing a model to additional tasks can help on the original task~\cite{ramsundar2015massively,odena2017conditional}.
Besides, an auxiliary classifier can be used for training the generator as well. 
Motivated by these considerations, we proposed a discriminator with an auxiliary classification module trained to reconstruct correct labels of adversarial examples and an improved loss function based on AC-GAN~\cite{odena2017conditional}.

In our approach, the discriminator has two branches as shown in Fig.~\ref{fig:overall_arc}: one is trained to discriminate between clean images \re{$x_{cln}$} and perturbed images $x_{pert}$; another is to classify $x_{pert}$ correctly.
\re{To further enhance the attack ability of the generator, we propose to train the classification module adversarially and add an attacker into the training process.
We choose PGD attack as our attacker here because of the strong attack ability.
Specifically, the attacker is used to generate adversarial examples for the classification module during training the discriminator.
Such a robust classification module would stimulate the generator to produce stronger attacks.
}
Another benefit of a robust discriminator is that it helps stabilize and accelerate the whole training~\cite{zhou2018dont}.

Overall, the loss function of our discriminator consists of three parts: $\mathcal{L}_S$ for discriminating clean/perturbed images and $\mathcal{L}_{C(adv)}$ for classification on adversarial examples generated respectively by the attacker and the generator, which are expressed as
\begin{equation}
\begin{array}{l}
\mathcal{L}_{S}=E\left[\log P\left(S=\textit {clean} \mid x_{\textit {clean}}\right)\right]+ \\
\qquad \qquad \qquad E\left[\log P\left(S=\textit {pert} \mid x_{\textit {pert}}\right)\right],
\end{array}
\end{equation}
\begin{equation}
\begin{array}{l}
\mathcal{L}_{C(adv)}=E\left[\log P\left(Class=y \mid x_{\textit {atk}}\right)\right],
\end{array}
\end{equation}
and
\begin{equation}
\begin{array}{l}
\mathcal{L}_{C(pert)}=E\left[\log P\left(Class=y \mid x_{\textit {pert}}\right)\right],
\end{array}
\end{equation}
where $y$ represents the true label.
The goal of the discriminator is to maximize $\mathcal{L}_{S}+\mathcal{L}_{C(adv)}+\mathcal{L}_{C(pert)}$.

\textbf{Generator.}
The generators in prior work~\cite{XiaoLZHLS18,DBLP:conf/cvpr/PoursaeedKGB18,mao2020gap++} are quite similar: given the clean images, the generators outputs the adversarial perturbations.
There are mainly two problems in these methods: 
1) their generators can generate only one specific targeted attack at a time because targeted classes are fixed during training;
2) their methods can hardly scale to large datasets.

To solve the above problems, we first modify the generator, which takes both clean images and targeted classes as inputs, as shown in Fig.~\ref{fig:overall_arc}. 
Then we propose to pre-train the encoder in a self-supervised way, which we elaborate on in the next subsection.
The pre-trained encoder can extract features effectively, and \re{reduce} the training difficulties from training scratch.
A pre-trained encoder's existence makes our approach similar to feature space attacks and \re{increases the adversarial examples' transferability~\cite{Inkawhich_2019_CVPR,zhou2018transferable}.}
As we train a robust discriminator with an auxiliary classifier, our generator's attack ability is further enhanced.

The loss function of the generator consists of three parts: $\mathcal{L}_{target(adv)}$ for attacking target models, $\mathcal{L}_{D(adv)}$ for attacking the discriminator, and $\mathcal{L}_{S}$ which is as same as the discriminator.
$\mathcal{L}_{target(adv)}$ and $\mathcal{L}_{D(adv)}$ are expressed as
\begin{equation}
\begin{array}{l}
\mathcal{L}_{target(pert)}=E\left[\log P_{target}\left(Class=t \mid x_{\textit {pert}}\right)\right],
\end{array}
\end{equation}
and 
\begin{equation}
\begin{array}{l}
\mathcal{L}_{D(pert)}=E\left[\log P_{D}\left(Class=t \mid x_{\textit {pert}}\right)\right],
\end{array}
\end{equation}
where $t$ is the class of targeted attacks.
The goal of the generator is to maximize $\mathcal{L}_{target(pert)}+\mathcal{L}_{D(pert)}-\mathcal{L}_{S}$.

The whole training procedure of AI-GAN can be found in Algorithm~\ref{algo:training}.

\begin{algorithm}[tbp]
\SetAlgoLined
\KwIn{Training data $\mathcal{D}_{train}$, targeted model $f$, learning rate $r$, the attacker $A$, class number $n$}
\KwOut{The parameters of the generator $G$ and the discriminator $D$ : $\theta_G$ and $\theta_D$.}
 initialization\;
 \For{each training iteration}{
    Sample $(x, y) \sim \mathcal{D}_{train}$\;
    Sample $t \sim \mathcal{U}_{n} $\;
    // training Generator
    $x_{pert} \leftarrow x + G(x)$\;
    $\mathcal{L}_{target(pert)} \leftarrow t^{\top} \log f(x_{pert})$\;
    $\mathcal{L}_{D(pert)} \leftarrow t^{\top} \log f(x_{pert})$\;
    $\mathcal{L}_{S} \leftarrow \mathbf{1}^{\top} \log f(x) + \mathbf{0}^{\top} \log f(x_{pert})$ \;
    $\theta_{G} \leftarrow \theta_{G} + r\nabla\left(\mathcal{L}_{target(pert)}+\mathcal{L}_{D(pert)}-\mathcal{L}_{S}\right) $\;
    // training Discriminator
    $x_{adv} \leftarrow x + A(x)$\; 
    $\mathcal{L}_{S} \leftarrow \mathbf{1}^{\top} \log f(x) + \mathbf{0}^{\top} \log f(x_{pert})$ \;
    $\mathcal{L}_{C(pert)} \leftarrow y^{\top} \log f(x_{pert})$\;
    $\mathcal{L}_{C(adv)} \leftarrow y^{\top} \log f(x_{adv})$\;
    $\theta_{D} \leftarrow \theta_{D} + r\nabla\left(\mathcal{L}_{C(adv)}+\mathcal{L}_{C(pert)} + \mathcal{L}_{S}\right) $\;
    
    }
 \caption{AI-GAN Training.}
 \label{algo:training}
\end{algorithm}

\subsection{Self-Supervised Pretraining of Genetator}
Both self-supervised learning and pretraining have proved to be effective in many vision or language problems~\cite{chen2020simple,misra2020self,radford2018improving,dong2019unified}.
Here we employ these two techniques to reduce the training difficulties and improve GAN's scalability in our problem.
The intuitive idea of self-supervised learning is to learn invariant features from data but without labels.
After self-supervised pretraining, the generator's encoder can compress data and extract useful features for attacks.
The loss we use for pretraining in our approach is the prevalent contrastive loss:
\begin{equation}
\mathcal{L}_{i, j}=-\log \frac{\exp \left(\operatorname{sim}\left(z_{i}, z_{j}\right) / \tau\right)}{\sum_{k=1}^{2 N}{[k \neq i]} \exp \left(\operatorname{sim}\left(z_{i}, z_{k}\right) / \tau\right)},
\end{equation}
where $z_i$ is the latent vector of sample $i$, $z_j$ is the latent vector of sample $j$, $\operatorname{sim}$ is a function to calculate the similarities of two latent vectors \eg $cos$, and \re{$\tau$ is a hyper-parameter. 
We follow the setting in~\cite{chen2020simple} and set $\tau$ as 0.5 in our experiments}.
\section{Experimental Results}\label{sec:experiment}
In this section, we conduct extensive experiments to evaluate \mbox{AI-GAN}.
\textbf{First}, we compare the attacking abilities of different methods with \mbox{AI-GAN} under different settings (white-box and black-box) on MNIST and CIFAR-10.
\textbf{Second}, we evaluate these methods with defended target models.
\textbf{Third}, we show the scalability of \mbox{AI-GAN} with on complicated datasets: CIFAR-100 and ImageNet-50.

\subsection{Datasets} 
We consider four different datasets in our experiments:
(1) MNIST~\cite{lecun-mnisthandwrittendigit-2010}, a handwritten digit dataset, has a training set of 60K examples and a test set of 10K examples.
(2) CIFAR-10~\cite{krizhevsky2009learning} contains 60K images in 10 classes, of which 50K for training and 10K for testing. 
(3) CIFAR-100. It is just like CIFAR-10, except that it has 100 classes containing 600 images each. 
(4) Imagenet-50, which is a randomly-sampled subset of ImageNet with 50 classes. 

In all of our experiments, \re{we follow the common settings in the literature~\cite{DBLP:conf/iclr/MadryMSTV18,DBLP:journals/corr/GoodfellowSS14,DBLP:journals/corr/SzegedyZSBEGF13,XiaoLZHLS18} and constrain the perturbations under a $L_{\infty}$ bound for different attack methods.
The $L_{\infty}$ bounds of perturbations are 0.3/1 on MNIST, and 8/255 on \mbox{CIFAR-10}, \mbox{CIFAR-100} and \mbox{ImageNet-50}.}

\subsection{Implementation Details} 
\subsubsection{Model Architectures and Training Details} 
We adopt generator and discriminator architectures similar to those of~\cite{miyato2018cgans} and use 7-step Projected Gradient Descent (PGD)~\cite{DBLP:conf/iclr/MadryMSTV18} as an attacker in the training process. 
We apply C\&W loss~\cite{carlini2017towards} to generating targeted adversarial examples for the Generator and set confidence $\kappa = 0$ in our experiments.
We use Adam as our solver~\cite{kingma2014adam}, with a batch size of 256 and a learning rate of 0.002.
A larger batch size is applicable as well with a larger learning rate.

\subsubsection{Target Models in the Experiments} 
For MNIST, we use model A from~\cite{DBLP:conf/iclr/TramerKPGBM18} and model B from~\cite{carlini2017towards}, whose architectures are shown in Table~\ref{tbl:models mnist};
For \mbox{CIFAR-10}, we use ResNet32 and WRN34 (short for Wide ResNet34)~\cite{he2016deep,zagoruyko2016wide};
For \mbox{CIFAR-100}, we use ResNet20 and ResNet32;
For \mbox{ImageNet-50}, we use ResNet18.
All the models are well pretrained on natural data. 

\begin{table}[htb]
\caption{Model architectures used in this work for MNIST dataset.}
\begin{center}
  \resizebox{\columnwidth}{!}{%
\begin{tabular}{cc}
  \toprule
  Model A & Model B \\
  \midrule
   & Conv(32, 3, 3)+ReLU \\
  Conv(64, 5, 5)+ReLU & Conv(32, 3, 3)+ReLU \\
  Conv(64, 5, 5)+ReLU & MaxPool(2, 2) \\
   Dropout(0.25) & Conv(64, 3, 3)+ReLU \\
   FullyConnected(128)+ReLU & Conv(64, 3, 3)+ReLU \\
   Dropout(0.5) & MaxPool(2, 2) \\
   FullyConnected + Softmax & FullyConnected(200) + ReLU \\
   & FullyConnected(200) + ReLU \\
   & Softmax \\
  \bottomrule
\end{tabular}}
\label{tbl:models mnist}
\end{center}
\end{table}

\subsubsection{Selection of Baselines} 
For comparison, we select three representative optimization-based methods: FGSM, C\&W attack, and PGD attack.
FGSM is a single-step attack method, which is fast but weak.
C\&W and PGD attacks are both iterative methods. 
Generally speaking, more steps will produce stronger attacks.
In our experiments, for C\&W attack, we follow the original setting stated in~\cite{carlini2017towards};
for PGD attack, we set steps as 20 and step size as 2/255.
We also compare with AdvGAN, the current state-of-the-art GAN-based attack method.

\subsection{White-box Attack Evaluation}
Attacking in white-box settings is the worst case for target models as the adversary knows everything about the models.
This subsection evaluates AI-GAN on MNIST and \mbox{CIFAR-10} with different target models.
The attack success rates of \mbox{AI-GAN} are shown in Table~\ref{table: semi-whitebox w/o d}.
From the table, we can see that \mbox{AI-GAN} achieves high attack success rates with different target classes on both MNIST and CIFAR-10.
On MNIST, the success rate exceeds 96\% given any targeted class.
The average attack success rates are 99.14\% for Model A and 98.50\% for Model B.
\mbox{AI-GAN} also achieves high attack success rates on \mbox{CIFAR-10}.
The average attack success rates are 95.39\% and 95.84\% for ResNet32 and 
WRN34 respectively. 
In this subsection, we mainly compared \mbox{AI-GAN} with AdvGAN, which is most similar to ours.
As shown in Table~\ref{table: comparison AI-GAN and AdvGAN}, \mbox{AI-GAN} performs better than AdvGAN in most cases.
It is worth noting that \mbox{AI-GAN} can launch different targeted attacks at once, which is superior to AdvGAN.

\begin{figure*}[htb]
  \subfloat[Adversarial examples and their perturbations on MNIST. ]{%
  \begin{minipage}{\textwidth}
  \includegraphics[width=.24\textwidth]{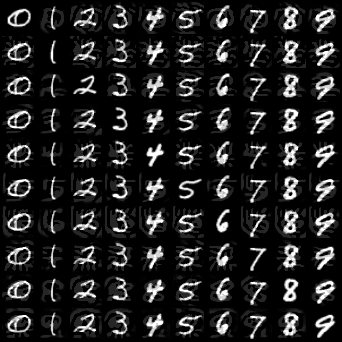}\hfill
  \includegraphics[width=.24\textwidth]{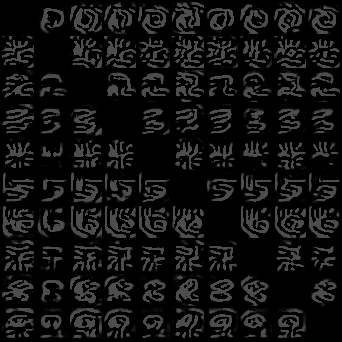}\hfill
  \includegraphics[width=.24\textwidth]{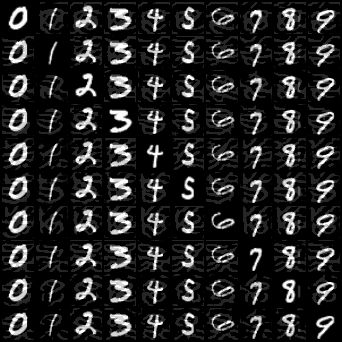}\hfill
  \includegraphics[width=.24\textwidth]{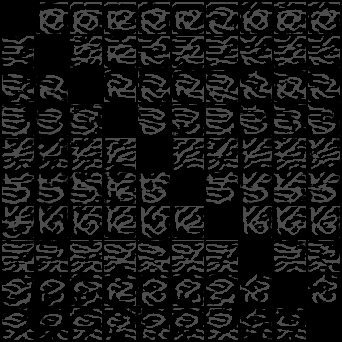}%
  \end{minipage}%
  \label{fig:wb mnist}
  }\par
  \subfloat[Adversarial examples and their perturbations on CIFAR-10.]{%
  \begin{minipage}{\textwidth}
  \includegraphics[width=.24\textwidth]{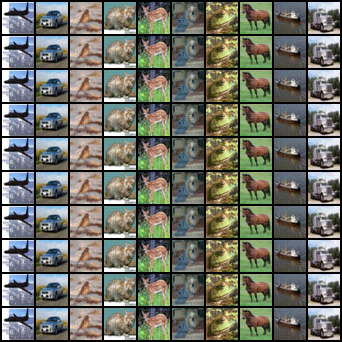}\hfill
  \includegraphics[width=.24\textwidth]{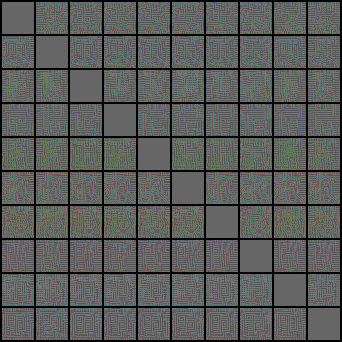}\hfill
  \includegraphics[width=.24\textwidth]{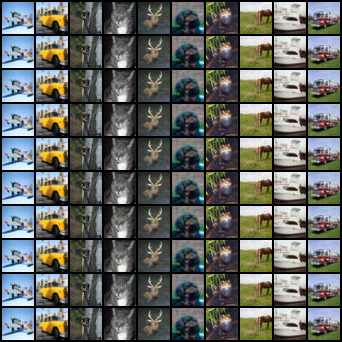}\hfill
  \includegraphics[width=.24\textwidth]{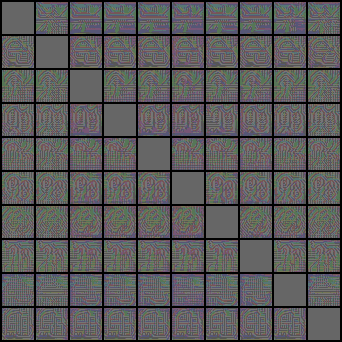}%
  \end{minipage}%
  \label{fig:wb cifar}
  }
  \caption{Visualization of Adversarial examples and perturbations generated by AI-GAN. Rows represent the different targeted classes and columns are 10 images from different classes. Original images are shown on the diagonal. Perturbations are amplified for visualization.}
   \label{fig:wb}
\end{figure*}

Randomly selected adversarial examples generated by AI-GAN are shown in Fig.~\ref{fig:wb}.
\begin{table}
\centering
\caption{Attack success rates of adversarial examples generated by AI-GAN against 
different different models on MNIST and \mbox{CIFAR-10} in white-box settings.}
\resizebox{\columnwidth}{!}{%
\begin{tabular}{ccccc}
\toprule
  & \multicolumn{2}{c}{MNIST} & \multicolumn{2}{c}{\mbox{CIFAR-10}} \\
  \cmidrule(r){2-3} \cmidrule(r){4-5}
  Target Class & Model A   &   Model B &  ResNet32 & WRN34 \\
  \midrule
Class 1      & 98.71\%     & 99.45\%     & 95.90\%        & 90.70\%     \\
Class 2      & 97.04\%     & 98.53\%     & 95.20\%        & 88.91\%     \\
Class 3      & 99.94\%     & 98.14\%     & 95.86\%        & 93.20\%     \\
Class 4      & 99.96\%     & 96.26\%     & 95.63\%        & 98.20\%     \\
Class 5      & 99.47\%     & 99.14\%     & 94.34\%        & 96.56\%     \\
Class 6      & 99.80\%     & 99.35\%     & 95.90\%        & 95.86\%     \\
Class 7      & 97.41\%     & 99.34\%     & 95.20\%        & 98.44\%     \\
Class 8      & 99.85\%     & 98.62\%     & 95.31\%        & 98.83\%     \\
Class 9      & 99.38\%     & 98.50\%     & 95.74\%        & 98.91\%     \\
Class 10     & 99.83\%     & 97.67\%     & 94.88\%        & 98.75\%     \\
\midrule
Average      & 99.14\%     & 98.50\%     & 95.39\%        & 95.84\%     \\
\bottomrule
\end{tabular}}
\label{table: semi-whitebox w/o d}
\end{table}  

\begin{table}
\caption{Comparison of attack success rate of adversarial examples generated by 
AI-GAN and AdvGAN in white-box setting.}
\label{table: comparison AI-GAN and AdvGAN}
\centering
\resizebox{\columnwidth}{!}{%
\begin{tabular}{ccccc}
\toprule
& \multicolumn{2}{c}{MNIST} & \multicolumn{2}{c}{\mbox{CIFAR-10}} \\
\cmidrule(r){2-3} \cmidrule(r){4-5} 
Methods & Model A   &   Model B &  ResNet32 & WRN34 \\
\midrule
AdvGAN & 97.90\% & 98.30\% & \textbf{99.30\%} & 94.70\% \\
AI-GAN & \textbf{99.14\%} & \textbf{98.50\%} & 95.39\% & \textbf{95.84\%} \\
\bottomrule
\end{tabular}}
\end{table}

\subsection{Black-box Attack Evaluation}
The black-box attack is another type of attack, which is more common in practice.
In this subsection, we evaluate \mbox{AI-GAN} to perform attacks in the black-box setting.
It is assumed that the adversaries have no prior knowledge of training data and the target models \eg parameters.
A common practice of the black-box attack is to train a local substitute model and apply the white-box attack strategy.
This is based on the transferability of adversarial examples.
We follow this practice and apply Dynamic Distillation to train AI-GAN.
Specifically, we construct a model $f_d$ and train it to distill information from target models.
Additionally, we jointly train it with AI-GAN.
Similar to adversarial training, $f_d$ is updated every time after updating AI-GAN.
Through $f_d$, AI-GAN can approximate the predictions of target models when fed with adversarial examples and learn to generate more aggressive adversarial examples.

In \re{the} implementation, We select Model C from~\cite{lecun1998gradient} and ResNet20 as the local models for MNIST and \mbox{CIFAR-10} respectively.
We mainly compare \mbox{AI-GAN} with AdvGAN as in the above subsection, and the experimental results are shown in \mbox{Table~\ref{tab: black box}}.
Apparently, adversarial examples generated by GAN-based methods have a better transferability than optimization-based methods.
\mbox{AI-GAN} and AdvGAN show their effectiveness and achieve higher attack success rates. 
On MNIST, \mbox{AI-GAN} achieves higher attack success rates than AdvGAN;
On CIFAR10, the success rates of \mbox{AI-GAN} is a little lower than AI-GAN, but it is acceptable considering the two methods' different training costs.

\begin{table}[htbp]
\caption{Comparison of attack success rate of adversarial examples generated by 
AI-GAN and AdvGAN in black-box setting.}
\label{tab: black box}
\resizebox{\columnwidth}{!}{%
\begin{tabular}{ccccc}
\toprule
\multicolumn{1}{c}{}       & \multicolumn{2}{c}{MNIST}                                 & \multicolumn{2}{c}{CIFAR 10}                               \\
\cmidrule(r){2-3} \cmidrule(r){4-5}
\multicolumn{1}{c}{Method} & \multicolumn{1}{c}{Model A} & \multicolumn{1}{c}{Model B} & \multicolumn{1}{c}{ResNet32} & \multicolumn{1}{c}{WRN34} \\
\midrule
FGSM                       & 29.49\%                     & 23.72\%                     & 20.86\%                       & 14.52\%                    \\
CW                         & 10.32\%                     & 10.09\%                     & 10.01\%                       & 10.07\%                    \\
PGD                        & 46.93\%                     & 33.70\%                     & 21.05\%                       & 14.78\%                    \\
AdvGAN                     & 93.40\%                     & 94.00\%                     & \textbf{81.80\%}              & \textbf{78.50\%}           \\
AI-GAN                     & \textbf{96.04\%}            & \textbf{94.95\%}            & 79.65\%                       & 75.94\%                   \\
\bottomrule
\end{tabular}%
}
\end{table}

\subsection{Attack Evaluation Under defenses}
\begin{table*}[htbp]
\caption{Comparison of attack success rates of adversarial examples generated by 
different methods in whitebox setting with defenses. 
\re{The best accuracy is bolded and the second best is underlined.}
}
\label{table: semi-whitebox w/ d}
\resizebox{\textwidth}{!}{%
\begin{tabular}{cccc|ccc|ccc|ccc}
\toprule
Dataset      & \multicolumn{6}{c}{MNIST}                                                                                       & \multicolumn{6}{c}{CIFAR10}                                                                          \\
\midrule
Target Model & \multicolumn{3}{c}{Model A}                            & \multicolumn{3}{c}{Model B}                            & \multicolumn{3}{c}{Resnet32}                        & \multicolumn{3}{c}{WRN34}                    \\
\cmidrule(r){2-4} \cmidrule(r){5-7} \cmidrule(r){8-10} \cmidrule(r){11-13}
Attacks      & Adv.             & Ens.             & Iter.Adv         & Adv.             & Ens.             & Iter.Adv         & Adv.            & Ens.             & Iter.Adv        & Adv.     & Ens.             & Iter.Adv        \\
\midrule
FGSM    & 4.30\%           & 1.60\%           & 4.40\%           & 2.70\%           & 1.60\%           & 1.60\%           & 5.76\%           & 10.09\%          & 1.98\%           & 0.10\%           & 3.00\%           & 1.00\%          \\
C\&W    & 4.60\%           & 4.20\%           & 3.00\%           & 3.00\%           & 2.20\%           & 1.90\%           & 8.35\%           & 9.79\%           & 0.02\%           & 8.74\%           & \textbf{12.93\%} & 0.00\%          \\
PGD     & {\ul 20.59\%}    & {\ul 11.45\%}    & \textbf{11.08\%} & 10.67\%          & 10.34\%          & 9.90\%           & 9.22\%           & {\ul 10.06\%}    & \textbf{11.41\%} & 8.09\%           & 9.92\%           & {\ul 9.87\%}    \\
AdvGAN  & 8.00\%           & 6.30\%           & 5.60\%           & {\ul 18.70\%}    & \textbf{13.50\%} & {\ul 12.60\%}    & \textbf{10.19\%} & 8.96\%           & 9.30\%           & {\ul 9.86\%}     & 9.07\%           & 8.99\%          \\
AI-GAN   & \textbf{23.85\%} & \textbf{12.17\%} & {\ul 10.90\%}    & \textbf{20.94\%} & {\ul 10.73\%}    & \textbf{13.12\%} & {\ul 9.85\%}     & \textbf{12.48\%} & {\ul 9.57\%}     & \textbf{10.17\%} & {\ul 11.32\%}    & \textbf{9.91\%}
 \\
\bottomrule
\end{tabular}%
}
\end{table*}

In this subsection, \re{we evaluate our method in the scenario, where the victims are aware of the potential attacks and defenses are deployed on target models.}
There are various defenses proposed against adversarial examples in the literature~\cite{DBLP:conf/iclr/MadryMSTV18,samangouei2018defensegan},
and adversarial training~\cite{DBLP:conf/iclr/MadryMSTV18} is widely accepted as the most effective way.
From these defense methods, we select three representative adversarial training methods to improve robustness of target models.
The first adversarial training method is proposed by~\cite{DBLP:journals/corr/GoodfellowSS14} based on Fast Gradient Sign Method (FGSM).
The objective function is expressed as $\tilde{J}(\boldsymbol{\theta}, \boldsymbol{x}, y)=\alpha J(\boldsymbol{\theta}, \boldsymbol{x}, y)+(1-\alpha) J\left(\boldsymbol{\theta}, \boldsymbol{x}+\epsilon \operatorname{sign}\left(\nabla_{\boldsymbol{x}} J(\boldsymbol{\theta}, \boldsymbol{x}, y)\right)\right.$.
The second method is Ensemble Adversarial Training extended from the first method by \cite{DBLP:conf/iclr/TramerKPGBM18}.
We use two different models as static models to generate adversarial examples.
Then these data is fed to the target model accompanying the clean data and adversarial data generated in each training loop.
Lastly, the third method is proposed by~\cite{DBLP:conf/iclr/MadryMSTV18}, which formulate adversarial training as a min-max problem and employs PGD to solve the inner maximization problem.
Note that, the adversaries don't know the defenses and will use the vanilla target models in white-box settings as their targets.

We compared \mbox{AI-GAN} with FGSM, C\&W attack, PGD attack and AdvGAN quantitatively under these defense methods, and the results are summarized in \mbox{Table~\ref{table: semi-whitebox w/ d}}. 
As we can see, \mbox{AI-GAN} has the highest attack success rates in most settings, and nearly outperforms all other approaches.

\subsection{Experiments on Real World Datasets}
\re{One concern of our approach is whether it can generalize to large datasets with more classes or high-resolution images.}
In this section, we demonstrate the effectiveness and scalability of our approach to CIFAR-100 and ImageNet-50.
We use these datasets because the current SOTA GAN~\cite{miyato2018cgans,Liu_2019_CVPR} work on these datasets, and adversarial training on ImageNet-1k data requires expensive computation resources \eg 128 GPUs~\cite{DBLP:conf/cvpr/XieWMYH19}, which is out of reach for us.

\begin{figure}[!htb] 
    \centering
  \subfloat[CIFAR-100\label{1a}]{%
      \includegraphics[width=0.5\columnwidth]{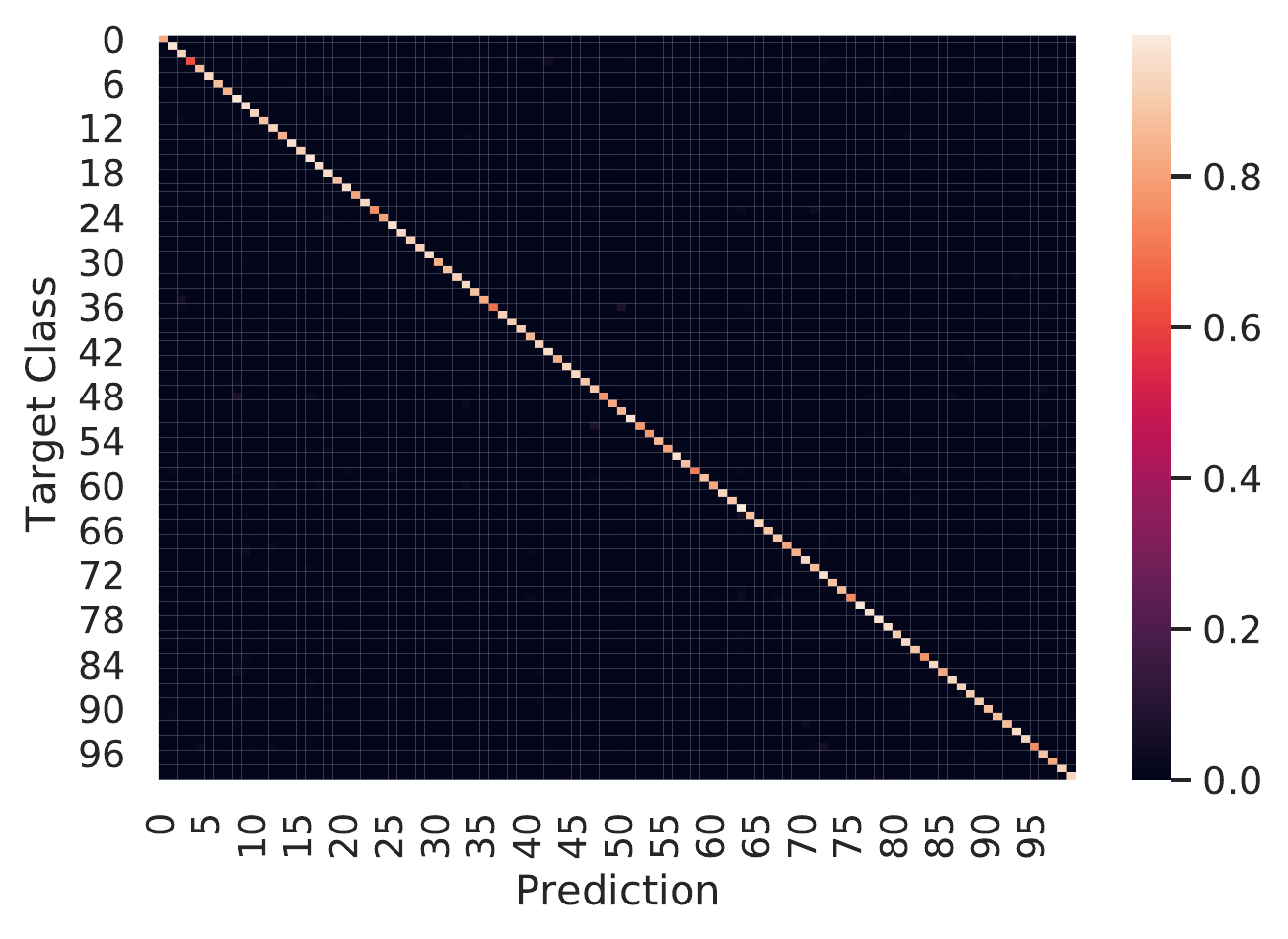}}
  \subfloat[ImageNet-50\label{1b}]{%
        \includegraphics[width=0.5\columnwidth]{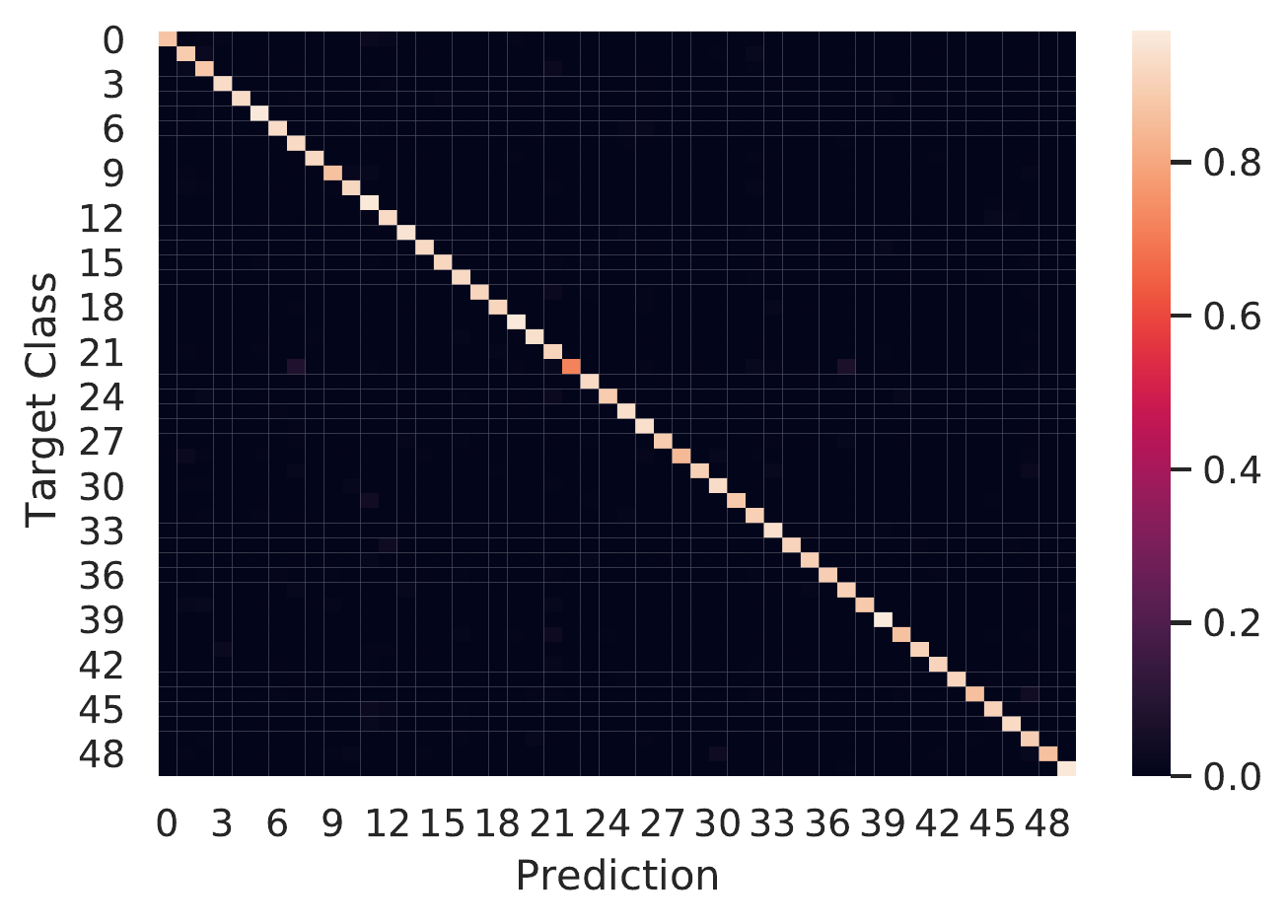}}
  \caption{Visualization of confusion matrix for targeted models on adversarial examples generated by AI-GAN, given data from CIFAR-100 and ImageNet-50 and different targeted classes. On the diagonal, it shows the targeted attacks success rates of AI-GAN. The lighter, the higher.}
  \label{fig: attack large dataset} 
\end{figure}

\begin{figure}[htbp]
\captionsetup[subfigure]{labelformat=empty}
\centering
\begin{subfigure}{.23\textwidth}
  \centering
  \includegraphics[width=\linewidth]{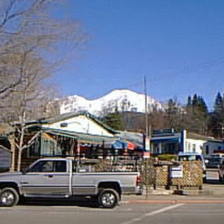}
  \caption{\re{\textbf{L}:\textit{truck}\quad \textbf{P}:\textit{barrier}}}
\end{subfigure}%
\hfill
\begin{subfigure}{.23\textwidth}
  \centering
  \includegraphics[width=\linewidth]{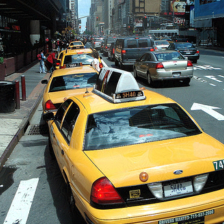}
  \caption{\re{\textbf{L}:\textit{car}\quad \textbf{P}:\textit{computer}}}
\end{subfigure}

\begin{subfigure}{.23\textwidth}
  \centering
  \includegraphics[width=\linewidth]{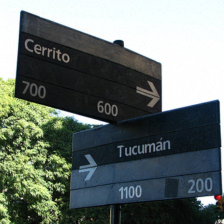}
  \caption{\re{\textbf{L}:\textit{street sign}\quad \textbf{P}:\textit{toy dog}}}
\end{subfigure}%
\hfill
\begin{subfigure}{.23\textwidth}
  \centering
  \includegraphics[width=\linewidth]{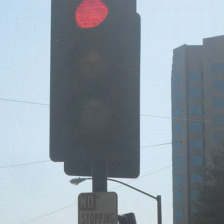}
  \caption{\re{\textbf{L}:\textit{traffic light}\quad \textbf{P}:\textit{truck}}}
\end{subfigure}

\caption{\re{Randomly sampled adversarial examples generated by AI-GAN on ImageNet-50. We use \textbf{L} and \textbf{P} to denote the \textbf{labels} and the \textbf{predictions} made by ResNet18, respectively.}}
\label{fig: sample large dataset}
\end{figure}

On CIFAR-100, we use ResNet20 and ResNet32 as our targeted models.
The average attack success rate of AI-GAN is 90.48\% on ResNet20, and 87.76\% on ResNet32.
On ImageNet-50, we use ResNet18 as our targeted model, and AI-GAN achieved a 92.9\% attack success rate on average.
All the attacks in our experiments are targeted, so we visualized the confusion matrix in Fig.~\ref{fig: attack large dataset}, where the rows are targeted classes, and columns are predicted classes.
The diagonal line in Fig.~\ref{fig: attack large dataset} shows the attack success rate for each targeted class.
We can see all the attack success rates are very high.
Samples of adversarial examples generated by AI-GAN on ImageNet-50 are shown in Fig.~\ref{fig: sample large dataset}.
Thus, our approach proves to be able to extend to complicated datasets with many classes or high-resolution images.

\begin{table}[!htb]
\caption{Comparison with the state-of-the-art attack methods.}
\label{tab:com_4_attacks}
\resizebox{\columnwidth}{!}{%
\begin{tabular}{cccccc}
\toprule
& FGSM  & C\&W             & PGD                      & AdvGAN           & AI-GAN           \\
\midrule
Run Time         & 0.06s & \textgreater{}3h & \multicolumn{1}{c}{0.7s} & \textless{}0.01s & \textless{}0.01s \\
Targeted Attack  & \checkmark     & \checkmark       & \checkmark          & \checkmark                & \checkmark                \\
Black-box Attack &       &   &        & \checkmark                & \checkmark               \\
\bottomrule
\end{tabular}%
}
\end{table}

\section{Discussions}\label{sec:disc}
We have demonstrated the strong ability of AI-GAN in Section~\ref{sec:experiment}.
In this section, we discuss the efficiency of the strong attack methods.

For evaluating the robustness of IoT systems in reality, efficiency is just as important as attack ability. 
To compare the efficiency of different methods, we perform 1000 attacks at a time and count the total time for each method, which are summarized in Table~\ref{tab:com_4_attacks}.

As we can see, GAN-based approaches have great advantages in saving time.
Even the fastest FGSM among optimization methods consumes multiple times of time, let alone PGD and C\&W attacks.
Moreover, FGSM is too weak for evaluation.
So there is always a trade-off between strong attack ability and efficiency, as it is usually true that optimization-based methods with more iterations are stronger.
In addition, these methods show poor black-box attack abilities in experiments, which is also supported by~\cite{XiaoLZHLS18}.

On the other hand, AI-GAN is superior to AdvGAN because AI-GAN can perform attacks with different targets once trained.
At the same time, AdvGAN needs extra copies with prefixed training targets, \ie ten AdvGANs needed for ten targeted classes.

\section{Conclusion}\label{sec:con}
In this paper, we propose \mbox{AI-GAN} to generate adversarial examples with different targets.
In our approach, a generator, a discriminator, and an attacker are involved in training.
Once AI-GAN is trained, it can launch adversarial attacks with different targets, which significantly  promotes efficiency and preserves image quality.
We compare AI-GAN with several SOTA methods under different settings \eg white-box, black-box or defended, and AI-GAN shows comparable or superior performances.
With the novel architecture and training objectives, AI-GAN scales to large datasets for the first time.
In extensive experiments, AI-GAN shows strong attack ability, efficiency, and scalability, making it a good tester for evaluating the robustness of IoT systems in practice.
For the future work, we would like to utilize the generative models to enhance the adversarial robustness of deep models~\cite{baluja2017adversarial,bai2021recent}.

\ifCLASSOPTIONcaptionsoff
  \newpage
\fi

\bibliographystyle{IEEEtran}
\bibliography{bibliography}

\end{document}